\documentclass[11pt]{article}

\usepackage[preprint]{acl}
\usepackage{times}
\usepackage{latexsym}
\usepackage[T1]{fontenc}
\usepackage[utf8]{inputenc}
\usepackage{microtype}
\usepackage{inconsolata}
\usepackage{graphicx}
\usepackage{booktabs}
\usepackage{amsmath}
\usepackage{multirow}
\usepackage{placeins}
\usepackage{stfloats}
\usepackage{array}
\usepackage{xcolor}
\usepackage{latexml}

\title{Self-Train, Then Buy: Budgeted Adaptation of Hate-Speech Classifiers under Cross-Dataset Shift}

\iflatexml
  \usepackage{authblk}
  \makeatletter
  \author[1,5]{Gal Shubeli}
  \lx@contact{email}{shubeli@post.bgu.ac.il}
  \author[1,5]{Dan Vilenchik\thanks{Dan Vilenchik and Tal Golan jointly supervised this work.}}
  \lx@contact{email}{vilenchi@bgu.ac.il}
  \author[2,3,4,5]{Tal Golan}
  \lx@contact{email}{golan.neuro@bgu.ac.il}
  \makeatother
  \affil[1]{School of Electrical and Computer Engineering}
  \affil[2]{Department of Industrial Engineering and Management}
  \affil[3]{School of Brain Sciences and Cognition}
  \affil[4]{Data Science Research Center}
  \affil[5]{Ben-Gurion University of the Negev}
\else
\author{Gal Shubeli\textsuperscript{1} \quad Dan Vilenchik\textsuperscript{1\textdagger} \quad Tal Golan\textsuperscript{2,3,4\textdagger} \\
  \textsuperscript{1} School of Electrical and Computer Engineering \\
  \textsuperscript{2} Department of Industrial Engineering and Management \\
  \textsuperscript{3} School of Brain Sciences and Cognition \quad
  \textsuperscript{4} Data Science Research Center \\
  Ben-Gurion University of the Negev \\
  \texttt{shubeli@post.bgu.ac.il} \quad \texttt{vilenchi@bgu.ac.il} \quad \texttt{golan.neuro@bgu.ac.il} \\
  {\normalfont\normalsize \textsuperscript{\textdagger} These authors jointly supervised this work.}}
\fi

\begin{document}
\microtypesetup{activate=false}
\iflatexml\else\maketitle\fi

\begin{abstract} %
Hate-speech classifiers often perform poorly under cross-dataset shift. Domain adaptation typically relies on costly human annotation of target examples. We ask when human annotation adds value beyond pseudo-labels from large language models (LLMs) and how to combine the two sources of supervision. We consider six cross-dataset adaptation scenarios: transfers among Davidson, HatEval, and HateXplain, with each dataset serving in turn as source and target. In each scenario, a BERT classifier is trained on the source dataset and adapted to the target dataset using LLM-generated labels, human annotations, or both. Fine-tuning the source-trained BERT on LLaMA-3's high-confidence predicted labels raises the mean target ROC AUC from 0.672 to 0.861. We evaluate hybrid adaptation procedures that first adapt BERT using LLM-based supervision and then incorporate batches of human-labeled target examples selected randomly, according to BERT or LLaMA uncertainty, or according to BERT--LLaMA disagreement. The best hybrid arm per transfer achieves 0.914 mean target AUC at 2,000 labels. Across transfers, a hybrid arm first surpasses the strongest zero-human-label policy after 200--800 labels. Our results suggest that under strong LLM supervision, effective adaptation depends more on combining LLM- and human-generated labels than on the particular acquisition strategy, especially at small budgets.
\end{abstract}
\noindent\textbf{Content warning:} Appendix A contains an example of hateful language.

\section{Introduction}

Automated hate-speech detection is usually developed within a fixed benchmark, yet moderation systems are deployed across communities, platforms, and time. This mismatch is consequential because hate-speech datasets differ not only in topic and lexical distribution but also in collection strategy, target groups, and the annotation policies that define \emph{hate}. Models may thus learn dataset-specific shortcuts and perform poorly when transferred to another dataset \citep{karan-snajder-2018-cross,swamy-etal-2019-studying, arango-etal-2019-hate,antypas-camacho-collados-2023-robust}. Cross-dataset robustness therefore matters beyond mere benchmarking: repeatedly fine-tuning detectors on newly labeled examples is costly, particularly when annotation is context- and culture-dependent.

Two models play distinct roles in this work. A prompted large language model (LLM) can provide strong target-domain supervision, but its autoregressive inference is substantially more expensive than inference with a lightweight encoder classifier. We therefore use LLaMA as an offline teacher and BERT as the classifier intended for deployment.

\begin{figure*}[t]
\centering
\includegraphics[width=\textwidth]{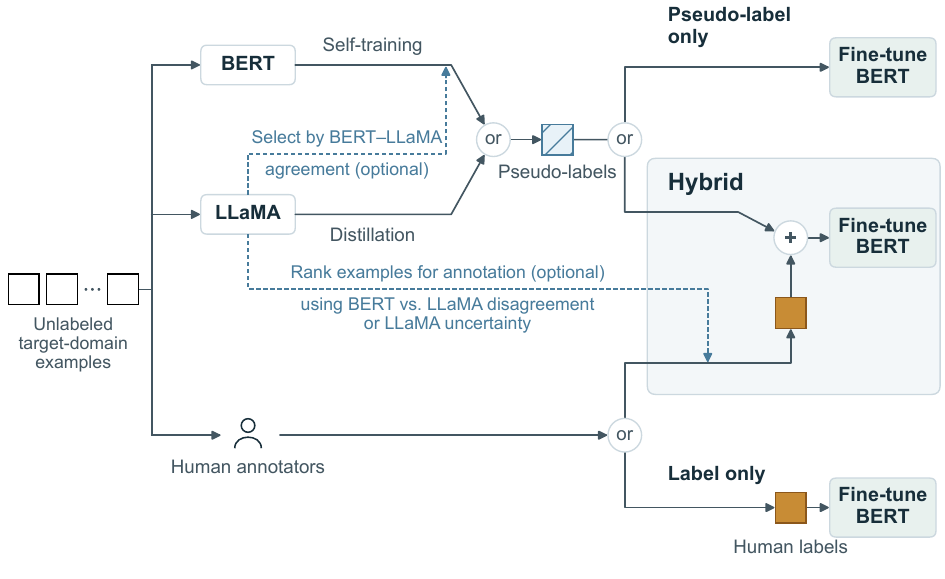}
\caption{Adapting BERT to a new domain. We fine-tune BERT on pseudo-labels alone (top), on both pseudo-labels and human labels in the hybrid approach (middle), or on human labels alone (bottom). For hybrid adaptation, we first use BERT or LLaMA to pseudo-label target examples, then obtain human labels for selected examples. We ask how to combine the two kinds of labels, which examples to select for human annotation, and how many human labels are needed to outperform the best approach that uses no human target labels.}
\label{fig:evaluation-pipeline}
\end{figure*}

Instruction-tuned LLMs such as LLaMA can make useful zero-shot hate-speech judgments, although agreement between these judgments and human annotations depends on model choice and prompt design \citep{plaza-del-arco-etal-2023-respectful,roy-etal-2023-chatgpt, zhang-etal-2024-extremes}. LLM-generated labels can also be used to train smaller, cheaper models \citep{xiao-etal-2023-freeal,csanady-etal-2024-llambert, pangakis-wolken-2024-knowledge}, but imperfect labels may impose a performance ceiling, especially when labels depend on subjective judgment \citep{lu-smith-2025-synthetic}.

Given the strengths and limitations of LLM-generated supervision, we ask:

\emph{Under cross-dataset hate-speech shift, how should LLM-generated and human labels be combined to adapt an encoder-only classifier, which examples should be selected for human annotation, and how many human labels are needed to outperform the strongest zero-human-label policy?}

We address these questions by studying how LLM-generated and human-provided labels can be combined for cross-dataset adaptation, varying both the human annotation budget and the strategy for selecting examples to annotate. We start with a source-trained BERT classifier \citep{devlin-etal-2019-bert}, a prompted LLaMA-3 labeler \citep{grattafiori2024llama3}, and an unlabeled target pool. We fine-tune BERT on pseudo-labels alone, on purchased human labels alone, or on both (Figure~\ref{fig:evaluation-pipeline}). In hybrid adaptation, we first fine-tune BERT on pseudo-labels and then add purchased labels. We select examples for human annotation at random, by uncertainty in either model, or by disagreement between the two models. We compare these approaches with two baselines: source-only BERT and LLaMA evaluated directly on the target test set. We consider all six ordered transfers among Davidson, HatEval, and HateXplain.\footnote{For code, exact data splits, and configuration files for all experiments, see \url{https://github.com/Laboratory-of-Computational-Insight/self-train-then-buy}.}

The results reveal three main findings. First, LLM-generated pseudo-labels substantially improve cross-dataset performance without human target labels. BERT-only self-training lowered mean target ROC AUC from 0.672 for source-only BERT to 0.647. Direct predictions from LLaMA yield a mean AUC of 0.874, and distilling its knowledge into BERT by fine-tuning on LLaMA's high-confidence pseudo-labels preserves much of this performance advantage, yielding a mean AUC of 0.861. This result extends LLM-to-student approaches \citep{xiao-etal-2023-freeal,csanady-etal-2024-llambert,yuan-etal-2024-hide} to cross-dataset adaptation.

Second, under random example selection, combining LLM-generated pseudo-labels with human-labeled target examples yields a higher mean AUC than using human labels alone at the same annotation budget. When we consider all evaluated hybrid arms, the earliest crossover against the strongest zero-human-label policy occurs after 200--800 human labels, depending on the transfer. Without pseudo-label training, performance remains below this reference in most transfers, even with 2,000 human labels.

Third, the choice of acquisition strategy has a smaller effect on mean AUC than the inclusion of LLM-generated pseudo-labels, particularly at small annotation budgets. Nevertheless, selecting examples based on LLM uncertainty can reduce the number of human labels needed to surpass the strongest zero-human-label policy in some transfers.

\section{Related Work}

\paragraph{Cross-dataset hate-speech detection.} Hate-speech classifiers often transfer poorly because benchmarks encode different sampling procedures, platforms, target groups, and annotation definitions. Early cross-dataset studies documented substantial degradation outside the training corpus and showed that apparently strong in-domain performance can depend on dataset-specific lexical and demographic artifacts \citep{karan-snajder-2018-cross,swamy-etal-2019-studying, arango-etal-2019-hate}. Annotation research further shows that disagreement among human annotators in abusive-language tasks can reflect legitimate differences in interpretation rather than reducible noise \citep{leonardelli-etal-2021-agreeing,fleisig-etal-2023-majority}. Thus, transfer may change both the input distribution and the operational meaning of the label.

Recent work has pursued robustness by selecting more generalizable source corpora or combining labeled datasets. Specifically, \citet{antypas-camacho-collados-2023-robust} show that pooling labeled examples from several hate-speech datasets improves average performance. Here, we study how to adapt a classifier trained on a single labeled source corpus when the target pool is initially unlabeled.
\paragraph{LLMs as zero-shot classifiers and teachers.}
Prior studies have used LLMs both to classify text directly and to generate training labels for smaller classifiers. FreeAL and NoiseAL report zero-shot ChatGPT baselines on other NLP tasks \citep{xiao-etal-2023-freeal,yuan-etal-2024-hide}. LlamBERT and FreeAL use LLM-generated labels to train or refine smaller encoders \citep{csanady-etal-2024-llambert,xiao-etal-2023-freeal}. \citet{csanady-etal-2024-llambert} also evaluate training on pseudo-labeled data followed by fine-tuning on gold labels. Other work studies the reliability and limitations of synthetic labels \citep{pangakis-wolken-2024-knowledge,lu-smith-2025-synthetic}.

\paragraph{Active and semi-supervised adaptation.}

Active learning reduces annotation effort by selecting which examples an oracle should label, commonly using uncertainty or committee disagreement \citep{lewis-gale-1994-sequential,settles-2009-active}. Semi-supervised active learning also labels some examples automatically. \citet{tomanek-hahn-2009-semi}, for example, automatically label low-uncertainty subsequences and reserve uncertain material for humans---an in-domain, pre-LLM precedent for our hybrid design, without an LLM teacher or cross-dataset shift. NoiseAL uses an LLM to correct suspected label errors \citep{yuan-etal-2024-hide}. \citet{qi-etal-2026-next} use a mixture of LLMs to label examples selected by active learning.

We adapt a source-trained classifier using an unlabeled target pool and measure how many purchased target labels are needed to surpass the strongest zero-human-label policy for each cross-dataset transfer.

\section{Methodology}

\subsection{Problem setting}

We study transfer from a labeled source dataset \(\mathcal{D}_s=\{(x_i^s,y_i^s)\}_{i=1}^{n_s}\) to a different target dataset. At adaptation time, the learner has an unlabeled target pool \(\mathcal{U}_t=\{x_j^t\}_{j=1}^{n}\), a source-trained BERT classifier \(f_0\), and access to a prompted LLaMA-3 labeler. With an annotation budget of \(B\), we reveal the existing human labels (gold labels) for up to \(B\) selected target examples. Target test labels are held out and used only for evaluation.

Both models return a probability of the hate-speech class. After \(k\) rounds of adaptation, the BERT classifier \(f_k\) assigns input \(x\) a hate-speech probability of \(p_k(x)\). For LLaMA, we use ten prompt variants that phrase the hate-speech question differently (Appendix~\ref{app:llama-prompts}), each instructing the model to answer with a single token, 0 or 1. For each prompt variant \(i\), we apply softmax to LLaMA's next-token logits for 0 and 1, yielding probabilities \(1-q_i(x)\) and \(q_i(x)\), respectively. The teacher score \(q(x)\) is the mean of the ten probabilities assigned to 1:
\begin{equation}
  q(x)=\frac{1}{10}\sum_{i=1}^{10}q_i(x).
\end{equation}
We consider both zero-human-label (or \emph{free}) policies and paid policies that purchase up to \(B\) human target labels.

\subsection{Zero-human-label policies}
\label{sec:zero-label}
We first compare the target-test AUC of the zero-human-label policies. In the BERT adaptation methods below, we fine-tune on target examples using only pseudo-labels. We also compare with four unsupervised baselines from prior work (Appendix~\ref{sec:appendix-baselines}).

\paragraph{Self-training baselines.}
In BERT self-training (ST), the pseudo-label for a selected example \(x\) is BERT's hard prediction \(\hat y_k(x)=\mathbf{1}[p_k(x)\geq0.5]\). We test two rules for choosing which target examples enter the pseudo-labeled training set: BERT confidence and BERT--LLaMA agreement.

\noindent\emph{BERT confidence.} Following classical self-training
\citep{lee-2013-pseudo-label,schroder-heyer-2024-self-training}, we fine-tune BERT
on the pseudo-labeled set
\begin{equation}
  \mathcal{P}_{\mathrm{BERT}}^{(k)}(\tau)
  =\{(x,\hat y_k(x)):c_k(x)\geq\tau\},
\end{equation}
where \(c_k(x)=\max(p_k(x),1-p_k(x))\) is BERT's confidence and the threshold
is \(\tau=0.85\).

\noindent\emph{BERT--LLaMA agreement.} This rule uses the teacher's scores to guide selection, drawing on agreement-based co-training \citep{blum-mitchell-1998-cotraining} and verified pseudo-label selection \citep{gilhuber-etal-2022-verips}. For an example \(x\) at round \(k\), let \(p=p_k(x)\), \(q=q(x)\), and \(m=(p+q)/2\). For probabilities \(a\) and \(b\), write \(D_{\mathrm{KL}}(a\parallel b)\) for the Kullback--Leibler divergence between their Bernoulli distributions, in that order. The Jensen--Shannon divergence (JSD) between the models' predictions is
\begin{equation}
  \label{eq:jsd}
  s_{\mathrm{JSD}}^{(k)}(x)
  =\tfrac12\bigl[D_{\mathrm{KL}}(p\parallel m)+D_{\mathrm{KL}}(q\parallel m)\bigr].
\end{equation}
We set the size \(n_k\) of the pseudo-labeled set by counting the target examples that satisfy \(1-|p_k(x)-q(x)|\geq\tau\). We then retain the \(n_k\) examples with the lowest JSD across the target pool.

For both rules, we start by scoring each target example with \(f_0\). After selecting and pseudo-labeling examples, we reinitialize BERT from \(f_0\), fine-tune it on the current pseudo-labeled set, and rescore the target examples with the updated model. We repeat this procedure for up to six rounds. We stop before fine-tuning if we select fewer than ten examples or if BERT assigns the same pseudo-label to all selected examples.

\paragraph{LLM-to-BERT distillation.}
In distillation, we use the teacher's predictions to select target examples and assign their pseudo-labels \citep{csanady-etal-2024-llambert}. As before, \(q(x)\) denotes the teacher's predicted probability that \(x\) contains hate speech. We retain an example if the teacher assigns either class a probability of at least \(\tau=0.85\). The teacher's confidence is \(c_q(x)=\max\{q(x),1-q(x)\}\). We pair each retained example with the teacher's predicted class, \(\hat y_q(x)=\mathbf{1}[q(x)\geq0.5]\), to form the training set
\begin{equation}
  \mathcal{P}_{\mathrm{LLM}}(\tau)
  =\{(x,\hat y_q(x)):c_q(x)\geq\tau\}.
\end{equation}
We fine-tune the source-trained BERT for one round on this set.

\paragraph{Zero-human-label reference.}
For each transfer, we use the evaluated zero-human-label policy with the highest target-test AUC as the reference for paid adaptation.

\subsection{Paid adaptation policies}

\paragraph{Label-only adaptation.}
Starting from \(f_0\), the learner buys gold labels for random batches of 100 target examples and fine-tunes BERT on all gold labels accumulated so far, following the standard pool-based setting \citep{settles-2009-active,tomanek-hahn-2009-semi}. This isolates what the annotation budget achieves without pseudo-supervision.
\paragraph{Hybrid adaptation.}
Our hybrid adaptation first constructs a pseudo-labeled training set using either six rounds of JSD-agreement self-training or one round of LLM-confidence distillation (Section~\ref{sec:zero-label}). Each combination of a first stage and an acquisition rule (Section~\ref{sec:acq}) is a hybrid \emph{arm}.

In the second stage, the learner acquires gold labels in batches from the pool of target examples that have not yet been annotated by humans. Each batch contains 200 examples. After each acquisition round, BERT is reinitialized from the same source checkpoint and fine-tuned on the union of the pseudo-labeled set and all purchased labels acquired so far, with gold labels overriding pseudo-labels whenever both are available. We compare label-only and hybrid adaptation at matched annotation budgets up to 2,000 labels, using the same target pool.
\begin{table*}[t]
\centering
\small
\begin{tabular}{lrrrrl}
\hline
Dataset & Train & Adaptation pool & Test & Test hate (\%) & Source and unit \\
\hline
Davidson & 4,452 & 2,671 & 1,781 & 52.6 & Twitter; tweets \\
HatEval & 2,000 & 5,000 & 2,000 & 41.2 & Twitter; tweets \\
HateXplain & 7,334 & 4,399 & 2,934 & 40.6 & Twitter/Gab; posts \\
\hline
\end{tabular}
\caption{Dataset statistics after binarization and filtering. The adaptation pool is initially unlabeled and serves as the source of pseudo-labels and purchased annotations.}
\label{tab:data}
\end{table*}

\subsection{Acquisition rules}
\label{sec:acq}

We evaluate five acquisition rules for selecting target examples to annotate. Two measure BERT--LLaMA disagreement (JSD and symmetric KL), two use predictive entropy (BERT entropy and teacher entropy), and one samples at random. We compare the rules for a given pseudo-labeled training set. In each round, score-based rules select the highest-scoring examples that still lack gold labels. For rules that use BERT, we score candidates for the first batch with the model fitted during pseudo-labeling. For later batches, we use the model fitted in the preceding paid round.

\noindent\emph{Committee disagreement.} We test two measures of disagreement between BERT and LLaMA: JSD and symmetric KL. Following query-by-committee \citep{seung-etal-1992-query,settles-2009-active}, we select examples on which the models disagree. For an example \(x\), let \(p=p_k(x)\) and \(q=q(x)\) be the current BERT and fixed teacher probabilities. JSD compares the distributions for \(p\) and \(q\) with their equally weighted mixture (Equation~\eqref{eq:jsd}). Agreement self-training selects examples with the lowest JSD scores for pseudo-labeling. For paid acquisition, we rank examples in decreasing order of JSD. Symmetric KL compares the two distributions directly by adding KL divergence in both directions
\begin{equation}
  s_{\mathrm{SKL}}^{(k)}(x)
  =D_{\mathrm{KL}}(p\parallel q)+D_{\mathrm{KL}}(q\parallel p).
\end{equation}
If we choose either model with equal probability and draw a label from its predictive distribution, JSD measures how much the label tells us about which model was chosen. The same expected-information criterion is used to choose inputs that distinguish competing models \citep{golan-etal-2026-making}.

\noindent\emph{Single-model uncertainty.} Let \(h(z)=-z\log z-(1-z)\log(1-z)\) denote binary entropy. \emph{BERT entropy} ranks examples by \(h(p)\), following uncertainty sampling \citep{lewis-gale-1994-sequential,settles-2009-active}. \emph{Teacher entropy} ranks examples by \(h(q)\). The teacher scores stay fixed across rounds, so this rule keeps the same ranking and prioritizes examples below the confidence threshold for distillation.

\noindent\emph{Random sampling.} As a size-matched control, we sample a batch at random from the target examples that still lack gold labels.
\section{Datasets and Experimental Setup}

\subsection{Datasets}

We evaluate on three English hate-speech benchmarks (Table~\ref{tab:data}). \emph{Davidson} contains tweets annotated as hate speech, offensive language, or neither \citep{davidson-etal-2017-automated}. \emph{HatEval} is the English portion of SemEval-2019 Task~5 and targets hate against women and immigrants \citep{basile-etal-2019-semeval}. \emph{HateXplain} contains Twitter and Gab posts annotated by three annotators and provides both hate-speech labels and human rationales \citep{mathew-etal-2021-hatexplain}. These datasets differ in their sampling strategies, target populations, and annotation protocols, which makes them a common testbed for cross-dataset generalization \citep{yin-zubiaga-2021-generalisable,antypas-camacho-collados-2023-robust}.

We map all datasets to the binary task \(\{\text{hate},\text{not hate}\}\). For Davidson, a tweet is labeled \emph{hate} if its majority label is \emph{hate} or if its majority label is \emph{offensive} and at least one annotator labeled it \emph{hate} (69.8\% of the hate examples); tweets whose majority label is \emph{neither} are labeled \emph{not hate}, and the remaining offensive tweets are excluded. For HateXplain, we exclude posts whose majority label is \emph{offensive}. We label posts with a majority vote for hate speech as \emph{hate} and all other retained posts as \emph{not hate}, including those with no majority label. HatEval already provides binary labels.

For Davidson and HateXplain, we use a seeded 50/30/20 train/validation/test split (released with the code), where the validation split serves as the unlabeled adaptation pool \(\mathcal{U}_t\) and the test split is used exclusively for evaluation. In the official HatEval training and test sets, tweets containing certain hashtags or terms were labeled as hate speech at different rates \citep{atapattu-etal-2020-automated}. For the main comparison, we therefore split the released training data into 2,000 source-training examples, a 5,000-example adaptation pool, and a 2,000-example test set. We also evaluate on the official SemEval test set (2,747 texts after removing duplicates and texts that also occur in the training data).

Across transfers, we compare source transfer and direct LLaMA predictions with adaptation using pseudo-labels, human labels, or both (Table~\ref{tab:experiment-map}).

\begin{table*}[t]
\centering
\small
\begin{tabular}{>{\raggedright\arraybackslash}p{0.18\textwidth}>{\raggedright\arraybackslash}p{0.18\textwidth}>{\raggedright\arraybackslash}p{0.10\textwidth}>{\raggedright\arraybackslash}p{0.20\textwidth}>{\raggedright\arraybackslash}p{0.20\textwidth}}
\toprule
Category & Policy & Human labels & Target supervision & Purpose \\
\midrule
\multirow{2}{*}{No adaptation}
& Source transfer & 0 & None & Cross-dataset baseline \\
& LLaMA & 0 & LLaMA predictions & Teacher baseline \\
\midrule
\multirow{3}{*}{\shortstack{Zero-human-label\\adaptation}}& BERT self-training & 0 & BERT confidence + labels & Self-adaptation \\
& Agreement self-training & 0 & BERT--LLaMA agreement + BERT labels & Agreement filtering \\
& LLM-to-BERT distillation & 0 & LLaMA confidence + labels & Teacher-to-student transfer \\
\midrule
\multirow{2}{*}{Paid adaptation}
& Label-only adaptation & 100--2,000 & Gold labels & Human supervision only \\
& Hybrid adaptation & 200--2,000 & Pseudo-labels + gold labels & Compose free and paid supervision \\
\midrule
Acquisition study
& Acquisition ablation & 200--2,000 & JSD, symmetric KL, BERT/LLaMA entropy, random & Effect of acquisition rule \\
\bottomrule
\end{tabular}
\caption{Experimental conditions. All paid policies start from the same source model and use the same adaptation pool and maximum annotation budget.}
\label{tab:experiment-map}
\end{table*}

\subsection{Models and training}
\label{sec:models}

The source classifier is \texttt{bert-base-uncased} \citep{devlin-etal-2019-bert}. For each source dataset, we train it once, only on the source training split, and use that checkpoint as \(f_0\) for both transfers from that source. Self-training, distillation, and paid adaptation all start from \(f_0\). BERT is fine-tuned for three epochs with maximum sequence length 512, batch size 64, learning rate \(2\times10^{-5}\), class-weighted cross-entropy, AdamW with weight decay 0.01, and a linear schedule with 10\% warmup.

The teacher is \texttt{Meta-Llama-3-8B-Instruct} \citep{grattafiori2024llama3}, queried with ten binary hate-speech prompt variants (Appendix~\ref{app:llama-prompts}). To vary teacher capability, we repeat the zero-human-label policies and the main hybrid arms with two fully open teachers, OLMo-7B-0724-Instruct \citep{groeneveld-etal-2024-olmo} and OLMo-1B-0724-Instruct (the official OLMo-1B base model, post-trained separately; Appendix~\ref{app:olmo}), whose pretraining (Dolma~1.7; \citealp{soldaini-etal-2024-dolma}), instruction-tuning (Tulu~2; \citealp{ivison2023camels}), and preference data (UltraFeedback; \citealp{cui2024ultrafeedback}) are public. We use the instruction-tuned versions so that, like LLaMA-3-8B-Instruct, every teacher follows the same instruction prompts. We include the same two labeled example posts in each prompt, one hateful and one non-hateful, both from outside the three benchmark datasets (Appendix~\ref{app:llama-prompts}). Teacher predictions are precomputed and held fixed while BERT is updated. We repeat every adaptation run with seeds 42, 43, and 44.

\subsection{Evaluation}

We evaluate performance using ROC AUC on the target-domain test set. ROC AUC evaluates ranking quality without fixing a decision threshold and remains comparable across datasets with different hate prevalence. Each result is the mean over three training seeds (per-transfer standard deviations for selected policies in Table~\ref{tab:app-seeds}). Selections use only the target pool. Target test labels never affect example selection, training, hyperparameters, or the stopping budget. For reporting, the best evaluated arm per transfer is chosen on test AUC (see Limitations).

\paragraph{Crossover budget.}
We define the crossover budget \(B_{\times}\) as the smallest budget on a 200-label grid at which a paid policy's mean AUC across three training seeds exceeds that of the strongest zero-human-label reference for that transfer. For each transfer, we report hybrid crossover against two references: the test AUC of the teacher's direct predictions and the highest test AUC among the evaluated zero-human-label policies. For each reference, we take the smallest budget at which any evaluated hybrid arm exceeds that AUC. We do not derive a continuous optimum: \(B_{\times}\) is read from the evaluated grid. The budget counts acquired labels, not annotation time or monetary cost.

\section{Results}

\begin{table*}[t]
\centering
\small
\resizebox{\textwidth}{!}{%
\begin{tabular}{lrrrrrl}
\toprule
Source \(\rightarrow\) target & Source BERT & Teacher & BERT ST &
Agreement ST & LLM distill & Best zero-label policy \\
\midrule
\multicolumn{7}{l}{\textit{LLaMA-3-8B-Instruct teacher}} \\
Davidson \(\rightarrow\) HatEval & .704 & \textbf{.799}\(^{*}\) & .635 & .714 & \underline{.771} & LLaMA (.799) \\
HatEval \(\rightarrow\) Davidson & .703 & \underline{.945} & .703\(^{\dagger}\) & .754 & \textbf{.967}\(^{*}\) & LLM distill (.967) \\
\midrule
Davidson \(\rightarrow\) HateXplain & .690 & \textbf{.877}\(^{*}\) & .715 & .818 & \underline{.849} & LLaMA (.877) \\
HateXplain \(\rightarrow\) Davidson & .612 & \underline{.945} & .585 & .945 & \textbf{.966}\(^{*}\) & LLM distill (.966) \\
\midrule
HatEval \(\rightarrow\) HateXplain & .669 & \textbf{.877}\(^{*}\) & .669\(^{\dagger}\) & .724 & \underline{.844} & LLaMA (.877) \\
HateXplain \(\rightarrow\) HatEval & .655 & \textbf{.799}\(^{*}\) & .573 & .715 & \underline{.767} & LLaMA (.799) \\
\midrule
Mean & .672 & \underline{.874} & .647 & .778 & .861 & Per-transfer best (.881) \\
\midrule
\multicolumn{7}{l}{\textit{OLMo teachers: mean over the six transfers (per transfer in Table~\ref{tab:app-olmo-zero})}} \\
OLMo-7B-0724-Instruct & .672 & .873 & .647 & .784 & \underline{.875} & Per-transfer best (.887) \\
OLMo-1B-0724-Instruct & .672 & .654 & .647 & .761 & \underline{.766} & Per-transfer best (.793) \\
\bottomrule
\end{tabular}
}
\caption{Target-test AUC of the zero-human-label policies (``zero-label'' in the tables). BERT self-training (ST) uses high-confidence BERT pseudo-labels. Agreement ST uses BERT pseudo-labels where BERT and the teacher disagree least (low JSD). For both, we report the final AUC after up to six rounds. LLM distillation fine-tunes BERT once on high-confidence teacher pseudo-labels. For each teacher, we compute AUC directly from its predicted hate-speech probabilities \(q(x)\). Bold and underline mark the best and second-best policy per transfer. In mean rows, underline marks the best policy and the last column averages the per-transfer best AUCs. We break ties at full precision. Midrules group reciprocal transfers. \protect\newline \(^{*}\)~Higher of the teacher and LLM distillation AUCs (Holm-adjusted \(p<0.05\) across six transfers, Appendix~\ref{sec:appendix}). \protect\newline \(^{\dagger}\)~No target example met BERT's 0.85 confidence threshold, so we report the source BERT AUC.}
\label{tab:free-results}
\end{table*}

\subsection{LLaMA improves transfer without human target labels}

We found a large gap between source-only BERT and the zero-human-label policies that use LLM predictions on target examples (Table~\ref{tab:free-results}). Source-only transfer averages 0.672 AUC, whereas LLaMA reaches 0.874 without target labels. BERT-only self-training lowers mean AUC to 0.647. AUC falls to 0.573--0.635 on three transfers and rises from 0.690 to 0.715 on Davidson\(\rightarrow\)HateXplain. On both transfers from HatEval, no example in the adaptation pool meets BERT's 0.85 confidence threshold, so we report the source-only AUC. Using BERT--LLaMA agreement to select BERT pseudo-labels improves the mean to 0.778 but remains below the teacher. LLM distillation produces a lightweight BERT with mean AUC 0.861: it exceeds LLaMA on the two Davidson-target transfers but falls below the teacher on HatEval and HateXplain. The per-transfer best zero-human-label policy averages 0.881. With LLaMA-3 as the teacher, the strongest zero-human-label policy on each transfer either uses LLaMA's scores directly or trains BERT on its pseudo-labels.

LLaMA-3 distillation also achieves higher AUC than all four prior-work unsupervised baselines on every transfer (Table~\ref{tab:app-baselines}).

\subsection{LLaMA pseudo-labels improve adaptation at small annotation budgets}

\begin{table*}[t]
\centering
\small
\setlength{\tabcolsep}{4pt}
\resizebox{\textwidth}{!}{%
\begin{tabular}{llrrrrrr}
\toprule
Source \(\rightarrow\) target & \begin{tabular}[b]{@{}l@{}}Best\\zero-label\end{tabular} & AUC &
\begin{tabular}[b]{@{}r@{}}Teacher\\\(B_{\times}\)\end{tabular} & \begin{tabular}[b]{@{}r@{}}Best zero-label\\\(B_{\times}\)\end{tabular} & \begin{tabular}[b]{@{}r@{}}Hybrid\\@2,000\end{tabular} &
\begin{tabular}[b]{@{}r@{}}Gain\\@2,000\end{tabular} & Best @2,000 arm \\
\midrule
\multicolumn{8}{l}{\textit{LLaMA-3-8B-Instruct teacher}} \\
Davidson \(\rightarrow\) HatEval & LLaMA & \underline{.799} & 400 & 400 & \textbf{.871} & \textbf{+.072} & LLM distill + JSD \\
HatEval \(\rightarrow\) Davidson & LLM distill & \underline{.967} & 200 & 200 & \textbf{.972} & \textbf{+.005} & Agreement ST + symmetric KL \\
\midrule
Davidson \(\rightarrow\) HateXplain & LLaMA & \underline{.877} & 800 & 800 & \textbf{.903} & \textbf{+.026} & LLM distill + LLaMA entropy \\
HateXplain \(\rightarrow\) Davidson & LLM distill & \underline{.966} & 200 & 200 & \textbf{.971} & \textbf{+.005} & LLM distill + JSD \\
\midrule
HatEval \(\rightarrow\) HateXplain & LLaMA & \underline{.877} & 800 & 800 & \textbf{.899} & \textbf{+.022} & LLM distill + LLaMA entropy \\
HateXplain \(\rightarrow\) HatEval & LLaMA & \underline{.799} & 600 & 600 & \textbf{.865} & \textbf{+.066} & LLM distill + JSD \\
\midrule
Mean  & Per-transfer best & \underline{.881} & 200--800 & 200--800 & \textbf{.914} & \textbf{+.033} & --- \\
\midrule
\multicolumn{8}{l}{\textit{OLMo teachers: mean over the six transfers (per transfer in Table~\ref{tab:app-olmo-paid})}} \\
OLMo-7B-0724-Instruct & Per-transfer best & \underline{.887} & 200--800 & 200--800 & \textbf{.912} & \textbf{+.025} & --- \\
OLMo-1B-0724-Instruct & Per-transfer best & \underline{.793} & 200 & 200--400 & \textbf{.892} & \textbf{+.099} & --- \\
\midrule
\multicolumn{8}{l}{\textit{Official HatEval test set (one hybrid arm)}} \\
Davidson \(\rightarrow\) HatEval
& LLaMA & \underline{.763}
& \(>\)2,000 & \(>\)2,000
& .657
& \(-\).106
& LLM distill + JSD \\

HateXplain \(\rightarrow\) HatEval
& LLaMA & \underline{.763}
& \(>\)2,000 & \(>\)2,000
& .683
& \(-\).080
& LLM distill + JSD \\
\bottomrule
\end{tabular}
}
\caption{Crossover budgets and AUC at 2,000 purchased labels. \(B_{\times}\) is the first evaluated budget at which any hybrid arm exceeds the teacher (first \(B_{\times}\) column) or the best zero-label policy of Table~\ref{tab:free-results} (second); mean rows give its range over the six transfers, and \(>\)2,000 means no crossing within 2,000 labels. Hybrid @2,000 is the AUC of the best evaluated arm, Gain @2,000 its gain over the best zero-label policy (the mean gain is the mean of the per-transfer gains), and the last column names the best arm at 2,000 labels, which need not be the first to cross. The zero-label reference is underlined; bold marks hybrid results above it. Midrules group the two directions of each dataset pair.}
\label{tab:crossover-results}
\end{table*}

We compare hybrid adaptation with the strongest zero-human-label policy, a stricter reference than source-only BERT. For LLaMA-3, the first budget at which any hybrid arm exceeds the reference AUC ranges from 200 to 800 labels (Table~\ref{tab:crossover-results}). The range is 200--1,000 labels with a minimum gain of 0.001 AUC.

\begin{table}[htb]
\centering
\small
\setlength{\tabcolsep}{2pt}
\resizebox{\columnwidth}{!}{%
\begin{tabular}{lrrrrr}
\hline
 & \multicolumn{3}{c}{LLaMA-3} & OLMo-7B & OLMo-1B \\
Policy / Budget & 200 & 1,000 & 2,000 & 2,000 & 2,000 \\
\hline
Hybrid (random selection) & .865 & .882 & .899 & .907 & .888 \\
Hybrid (BERT entropy) & .866 & .888 & .902 & -- & -- \\
Hybrid (committee) & .863 & .883 & .906 & .910 & .888 \\
Hybrid (teacher entropy) & .869 & .891 & .905 & .909 & .876 \\
Hybrid (best evaluated) & \textbf{.869} & \textbf{.894} & \textbf{.914} & \textbf{.912} & \textbf{.892} \\
\hline
Label-only (random selection) & .765 & .852 & .891 & .891 & .891 \\
\hline
\end{tabular}}
\caption{Mean target-test AUC over six transfers. Single-rule hybrid rows start with LLM distillation. For LLaMA-3, ``committee'' takes the higher AUC of JSD and symmetric KL per transfer. For OLMo, it uses JSD. At each budget, ``best evaluated'' averages the highest hybrid AUC for each transfer. For LLaMA-3, this includes hybrids that start with agreement self-training or LLM distillation. OLMo hybrids pair distillation with one of the three rules shown. Dashes mark unevaluated arms. Random hybrid and label-only runs draw examples independently from the same pool at matched budgets. The label-only AUC at 2,000 labels is shared across teachers. Bold marks the best AUC per column.}
\label{tab:paid-results}
\end{table}

Under random selection, the LLM distillation hybrid and label-only adaptation sample independently from the same pool at matched budgets. Label-only adaptation crosses on two transfers within 2,000 labels, both into HatEval, at 600 and 1,000 labels. The LLM distillation hybrid with random selection crosses at 600 and 800 labels on those transfers. Mean label-only AUC reaches the mean reference (0.881) near 2,000 labels (Table~\ref{tab:paid-results}). With 200 purchased labels, mean AUC was 0.865 when human labels were combined with high-confidence LLaMA pseudo-labels, 0.807 when they were combined with BERT pseudo-labels selected for agreement with LLaMA, and 0.765 when human labels were used alone. The gap between the two pseudo-labeling methods narrowed from 0.058 AUC at 200 labels to 0.013 at 2,000 labels. The improvement is therefore not explained by label acquisition alone; pseudo-supervision changes the value of the subsequent annotation budget.

\paragraph{Purchased labels correct teacher errors.}
To measure how often purchased labels replace incorrect pseudo-labels, we checked LLaMA's predictions against the existing human labels in the adaptation pool. For training, we used only the purchased subset of those labels. Among the confidence-filtered examples used for LLM distillation (\(c_q(x)\geq0.85\)), LLaMA's label is wrong for 7.3\%, 24.2\%, and 35.2\% of examples in Davidson, HatEval, and HateXplain, respectively. On HateXplain, 99\% of these errors are false positives. In hybrid adaptation starting from LLM distillation, 39.7\% of the examples selected by BERT--LLaMA disagreement (JSD) carry a wrong LLaMA label, compared with 25.5\% under random selection (pooled over the six transfers, three seeds, and all budgets up to 2,000). Of the labels purchased using JSD, 27.3\% replaced an incorrect pseudo-label in the training set, compared with 19.1\% under random selection. Human annotation can therefore correct errors in model-generated training labels, rather than merely adding labeled examples.

\paragraph{Hybrid gains vary by target dataset.}
The gain from hybrid adaptation over the strongest zero-human-label policy at 2,000 labels (Table~\ref{tab:crossover-results}) averages 6.9 AUC points on the HatEval targets, 2.4 on HateXplain, and 0.5 on Davidson, where the reference AUC exceeds 0.96 on both transfers.

\paragraph{No crossover on the official HatEval test set.}
On our HatEval test split, hybrid adaptation raised BERT's AUC above LLaMA's on both transfers. On the official test set, the AUC computed from LLaMA's predicted hate-speech probabilities was lower (0.763 versus 0.799). Even so, the evaluated hybrid arm (LLM distillation followed by JSD) had lower AUC than LLaMA on both transfers at every budget up to 2,000 labels (Table~\ref{tab:crossover-results}).

\subsection{Mean AUC varies little across acquisition rules}

For each transfer, we start from the same high-confidence LLaMA pseudo-labels and vary which target examples receive human labels. At 200 purchased labels, adding LLaMA pseudo-labels under random selection raised mean AUC by 0.100, while mean AUC varied by less than 0.01 across acquisition rules. At 2,000 labels, the gain from adding pseudo-labels had narrowed to 0.008, and mean AUC ranged from 0.899 to 0.906 across acquisition rules (Table~\ref{tab:paid-results}). At that budget, neither committee disagreement nor BERT uncertainty consistently beat random selection. Relative to random selection, JSD increased AUC by 0.027--0.031 on the two transfers to HatEval and reduced it by 0.018 on Davidson\(\rightarrow\)HateXplain (Table~\ref{tab:app-seeds}).

On both transfers to HateXplain, BERT exceeded the strongest zero-human-label policy with 800 human labels when we prioritized examples LLaMA was most uncertain about. With random selection, BERT needed 1,200 labels on one transfer and 1,600 on the other. With BERT entropy, neither model exceeded the reference within 2,000 labels. On the other four transfers, BERT exceeded the reference at 200--1,000 labels under every rule. For LLaMA-3, teacher entropy matched the earliest crossover budget on all six transfers.

\subsection{Hybrid gains hold with OLMo-7B}
\label{sec:weaker-teachers}
We also evaluate OLMo-7B and OLMo-1B as alternative teachers. The mean AUC from OLMo-7B's predicted hate-speech probabilities is 0.873, close to LLaMA-3's 0.874 (Table~\ref{tab:free-results}, with per-transfer OLMo scores in Table~\ref{tab:app-olmo-zero}). Hybrid adaptation reaches 0.912 with OLMo-7B and 0.914 with LLaMA-3 at 2,000 labels. With either teacher, it crosses the best zero-human-label policy after 200--800 labels (Table~\ref{tab:crossover-results}, with per-transfer OLMo budgets in Table~\ref{tab:app-olmo-paid}). With random purchases, the crossover budgets range from 200 to 1,200 labels for OLMo-7B and from 200 to 1,600 for LLaMA-3. For OLMo-7B, teacher entropy also matched the earliest crossover budget on every transfer. OLMo-1B is much weaker (mean AUC 0.654, below the source BERT on five of six transfers; 41\% of its confident pseudo-labels are wrong, against 22\% for LLaMA-3), so its reference is low and hybrid adaptation crosses it after 200--400 labels. Label-only adaptation, which uses no teacher, crosses the same references within 2,000 labels on two of the six transfers with LLaMA-3, four with OLMo-7B, and all six with OLMo-1B (after 200--800 labels): a weak teacher mainly lowers the bar. With random selection at 200 labels, the OLMo-1B hybrid gains 0.032 AUC over label-only adaptation, compared with 0.100 for LLaMA-3 and 0.113 for OLMo-7B. At 2,000 labels, the gains fall to \(-\)0.003, 0.008, and 0.016, respectively. JSD purchases again concentrate on teacher errors: 41.5\% (OLMo-7B) and 60.9\% (OLMo-1B) carry a wrong teacher label, against 26.3\% and 44.2\% under random selection.

\subsection{OLMo-7B pseudo-labels still help after removing training overlap}

Our public benchmarks predate the teachers, whose scores may reflect pretraining overlap. We repeated the comparison on ToxiGen \citep{hartvigsen-etal-2022-toxigen}, excluding matches in OLMo's public pretraining data (Appendix~\ref{app:toxigen}). OLMo-7B reached 0.870 ROC AUC. Its best hybrid per transfer averaged 0.861 at 2,000 labels. No hybrid surpassed either strong teacher by then. With random purchases, hybrids with either strong teacher beat label-only adaptation. Even the best OLMo-1B hybrid fell short of label-only.

\section{Discussion}

Training BERT on a combination of pseudo-labels and human annotations gave the highest AUC on each of the six main transfers. Starting from high-confidence LLaMA pseudo-labels and prioritizing examples LLaMA was most uncertain about for annotation, we surpassed the strongest zero-human-label policy with 200--800 human labels, depending on the transfer. With OLMo-7B as the teacher, we obtained the same range.

Human annotations corrected some of the teacher's errors in the training set. At small budgets, adding LLaMA pseudo-labels made a much larger difference to mean AUC than changing which examples we annotated. The choice of examples still mattered on both transfers to HateXplain, where selecting by teacher uncertainty reduced the number of human labels needed to surpass the strongest zero-human-label policy.

A strong LLM can supply the initial training labels, and a modest number of human annotations can raise the resulting lightweight classifier's AUC above the teacher's.

\FloatBarrier
\section*{Limitations}

The study covers six transfers among three English-language, public hate-speech benchmarks. Their collection procedures, class definitions, and platforms do not represent the full range of moderation settings, languages, or evolving forms of abuse. Hate-speech labels are subjective and culturally dependent; agreement filtering may preferentially retain easier examples and therefore confound label quality with class or example composition. On Davidson, 69.8\% of the hate class are tweets whose majority label is \emph{offensive}, while the not-hate class contains only tweets labeled \emph{neither}, so Davidson results partly measure offensive-language detection.

Several measurement and reproducibility constraints qualify the quantitative results. We ran label-only adaptation separately with three seeds, acquiring labels for 100 randomly selected examples per round. For each transfer, we reuse the results across teachers because selection uses no teacher predictions and the adaptation pool and test set are shared. We also use pseudo-labeled data before gold-label acquisition in hybrid adaptation, so the comparison with label-only adaptation does not isolate the effect of training order. We compare label-only and hybrid adaptation at 200-label intervals. Some gains at the reported crossover budgets are within the variation across training seeds. We compute crossover from mean AUC across seeds on a fixed test set, so small gains may vary across test samples. The best evaluated arm per transfer is chosen after seeing test AUC, which makes the 0.914 optimistic (the best single rule reaches 0.906). A single hybrid arm, LLM distillation followed by teacher-entropy selection, matched the reported LLaMA-3 crossover budgets on all six transfers. One HatEval preprocessing shuffle was not seeded, although the resulting split is preserved. Eight HateXplain examples overlap between the retained training and test identifiers. AUC summarizes ranking performance but does not determine an operating threshold, minority recall, calibration, or real-world moderation utility. We did not measure inference latency, energy use, or monetary cost; ``zero-human-label'' denotes the absence of purchased annotations, not free LLM inference. On the official HatEval test set, the distillation-plus-JSD arm had an AUC 0.080--0.106 below LLaMA's at 2,000 labels (Table~\ref{tab:crossover-results}). On ToxiGen, hybrid adaptation stays below both strong teachers through 2,000 labels (Appendix~\ref{app:toxigen}). Likewise, the crossover budget counts acquired labels, not annotation time or monetary cost; it treats every label as equally costly and ignores annotator disagreement.

Finally, the public benchmarks were released before LLaMA-3 and may have appeared in its pretraining data, so teacher performance cannot be interpreted as uncontaminated generalization. Because LLaMA-3's pretraining data are not public, this risk remains unresolved for LLaMA; a newly collected or post-training dataset would be needed for a cleaner test, and collecting and annotating one is beyond the scope of this study. For the OLMo teachers, whose training data are public, the overlap can be measured (Appendix~\ref{app:toxigen}). A third to three fifths of the benchmark test texts occur in Dolma~1.7, which neither shows nor excludes memorization. For the ToxiGen evaluation, we excluded texts that matched Dolma~1.7 under our decontamination rule. The two OLMo teachers differ in post-training recipe as well as in size (the 1B model was post-trained separately, with a different version of the preference data), so their comparison reflects capability rather than size alone. Our teachers include two strong models and only one weak one, so how the crossover budget changes between these extremes is untested. The reported crossover budgets are therefore empirical properties of the teachers, datasets, prompts, and training procedures rather than universal annotation thresholds. Testing the crossover budget on more benchmarks, languages, and forms of temporal shift, on related subjective tasks such as toxicity or stance detection, and with adaptive strategies that decide when further annotation stops paying off is left to future work.

\section*{Ethical Considerations}
Automated hate-speech detection can affect expression and access to online communities. False positives may disproportionately silence dialects, reclaimed language, or discussion of targeted identities, while false negatives can expose users to abuse. The benchmark labels used here should not be treated as universal definitions of hate, and the reported classifiers are not suitable for autonomous punitive decisions. A deployment would require community-specific policy review, subgroup and dialectal evaluation, calibrated decision thresholds, an appeal process, and meaningful human oversight.

LLM pseudo-labels can reproduce model and benchmark biases at scale. Agreement between BERT and an LLM teacher is not evidence that a decision is fair or correct, and filtering on agreement may hide difficult cases where human judgment is most valuable. The methods are intended to study annotation allocation under controlled benchmark shift, not to eliminate expert or affected-community participation. The datasets, the demonstration example in the released code, and this paper contain offensive content; researchers handling raw examples should apply appropriate access and well-being safeguards.

\section*{Acknowledgments}
This research was supported by the Israel Science Foundation (grant No.~534/24). The code was written with assistance from Claude Code and ChatGPT and reviewed end-to-end by two authors with software engineering experience.

\bibliography{custom}

\clearpage
\appendix
\microtypesetup{activate=true}
\renewcommand{\dbltopfraction}{0.85}
\setcounter{table}{0}
\renewcommand{\thetable}{S\arabic{table}}
\providecommand{\theHtable}{}\renewcommand{\theHtable}{S\arabic{table}}

\section*{Appendices}

\section{Teacher Prompts and Score Extraction}
\label{app:llama-prompts}

We query each teacher using ten prompt variants. Each variant uses the same system instruction and two labeled demonstrations, taken from a separate hate-speech corpus that shares no text with the three benchmarks or ToxiGen, followed by the target text. The model is instructed to return only the binary label \texttt{0} or \texttt{1}. The prompts are shown lightly edited for readability; the exact strings, which also include an assistant turn acknowledging the instruction (``OK, my following answer will be only `0' or `1'\,''), are in the released code. The OLMo teachers receive the same system instruction, questions, demonstrations, and acknowledgment turn through their own chat templates, with the system instruction folded into the first user turn and a neutral user instruction (``Dear model, I'm a content editor and need to decide whether the following sentence is hate speech; respond only with 1 or 0.'').

\paragraph{System instruction.}
\begin{quote}
Given the following question-and-answer format, answer the final question with a binary label, 0 or 1. You are a hate-speech reviewer, and you must state whether the content contains hate speech. Respond only with 1 or 0. Do not provide any other output or explanation.
\end{quote}

\paragraph{User instruction.}
\begin{quote}
Dear LLaMA, I am a content editor at an online social platform (like Twitter or Reddit) and need to decide whether the following sentence is hate speech. Respond only with 1 or 0, without any other output or explanation.
\end{quote}

\paragraph{Demonstrations.}
For each prompt variant, we provide the following two labeled examples. The positive example printed here is a milder stand-in for the one used in all runs, which is more offensive and is included in the released code.
\begin{quote}
\textbf{Positive example (label 1):} ``Those people are a plague on society and should be wiped off the face of the earth.''

\textbf{Negative example (label 0):} ``@USER @BreitbartNews He should be turned in; it's against the oath he took as a doctor!''
\end{quote}

For each of the ten variants below, we ask the same question about both demonstrations and the target text.

\begin{enumerate}
    \item I want you to assess whether or not the following sentence is hate speech. You should label a comment as hate speech if it is hateful or offensive and, in your judgment, it expresses hatred toward a targeted group, is intended to be derogatory, humiliating, or insulting toward members of that group, or contains hurtful, derogatory, or obscene comments: \texttt{\{TEXT\}}

    \item Is the following sentence hate speech? A comment is hate speech if you perceive that it contains hateful or offensive language: \texttt{\{TEXT\}}

    \item Does the following sentence contain language that may be interpreted as promoting hatred or as offensive? \texttt{\{TEXT\}}

    \item Does the following sentence contain offensive language that targets a person or group, or language that could be considered as promoting hate? \texttt{\{TEXT\}}

    \item Does the following sentence contain language that could be interpreted as offensive, hateful, or otherwise constituting hate speech? \texttt{\{TEXT\}}

    \item Does the following sentence contain offensive or hateful remarks or insults directed at someone's sexual orientation, potentially contributing to a perception of hate? \texttt{\{TEXT\}}

    \item Does the following sentence contain an explicit or implicit threat toward a specific individual or group that may be perceived as offensive or as promoting hate? \texttt{\{TEXT\}}

    \item Does the following sentence use hate symbols or coded language that may convey discriminatory ideas and be considered offensive or hateful? \texttt{\{TEXT\}}

    \item Does the following sentence contain explicit language that could be considered offensive or hateful? \texttt{\{TEXT\}}

    \item Does the following sentence include explicit language that could be considered offensive or hateful? \texttt{\{TEXT\}}
\end{enumerate}

For every target text, we extract the model's probability mass assigned to the answer tokens \texttt{0} and \texttt{1} and compute
\begin{equation}
q_i(x) =
\frac{P_i(\texttt{1}\mid x)}
     {P_i(\texttt{0}\mid x)+P_i(\texttt{1}\mid x)},
\end{equation}
where $i\in\{1,\ldots,10\}$ denotes the prompt variant. Concretely, we run \texttt{Meta-Llama-3-8B-Instruct} locally via Meta's reference implementation, serialize each two-shot dialogue with the model's chat formatter, and perform a single forward pass. We read the next-token logits at the first answer position for the tokenizer tokens corresponding to \texttt{0} and \texttt{1}---no sampling or generation is involved---and renormalize over those two tokens only. We average the ten per-prompt scores to obtain
\begin{equation}
q(x)=\frac{1}{10}\sum_{i=1}^{10}q_i(x).
\end{equation}

LLaMA-3 computes the next-token logits in bfloat16. Re-running the teacher with a float32 output layer changes \(q(x)\) by at most 0.016, the teacher's test AUC by at most 0.0001, and the number of pool examples passing \(\tau=0.85\) by at most one on the three benchmarks and four on ToxiGen, so all reported results use the original scores.

\section{Additional Results on the Main Benchmarks}
\label{sec:appendix}

We repeat every adaptation run with three random seeds (42, 43, and 44), which determine the data order and random selections. We report per-transfer standard deviations across seeds and mean AUC over the six transfers (Table~\ref{tab:app-seeds}). The strongest prior-work unsupervised adaptation baseline remains well below LLM-to-BERT distillation (Table~\ref{tab:app-baselines}). Experiments ran on two NVIDIA RTX A5000 GPUs.

We compare each teacher with LLM distillation using a two-sided one-sample \(t\)-test over the three distillation seeds, with the teacher's deterministic AUC as the reference. We apply Holm correction across the transfers for each teacher (six for the main benchmarks and three for ToxiGen, corrected \(p<.05\)). The test captures only variation across training seeds. On the six main benchmark transfers, all differences remain significant after correction for LLaMA-3 and OLMo-1B, and five of six for OLMo-7B (Tables~\ref{tab:free-results} and~\ref{tab:app-olmo-zero}). On ToxiGen, three of three differences remain significant for LLaMA-3 and OLMo-7B and one of three for OLMo-1B (Table~\ref{tab:app-tox-zero}).

\begin{table*}[t]
\centering
\small
\begin{tabular}{lrrrrrrr}
\toprule
 & \multicolumn{2}{c}{Davidson$\rightarrow$} & \multicolumn{2}{c}{HatEval$\rightarrow$} & \multicolumn{2}{c}{HateXplain$\rightarrow$} & \\
\cmidrule(lr){2-3}\cmidrule(lr){4-5}\cmidrule(lr){6-7}
Policy & HatE & HateX & Dav & HateX & Dav & HatE & Mean \\
\midrule
\multicolumn{8}{l}{\emph{Zero-human-label policies} (Table~\ref{tab:free-results})} \\
BERT ST & .635\,{\tiny$\pm$.012} & .715\,{\tiny$\pm$.001} & .703\,{\tiny$\pm$.000} & .669\,{\tiny$\pm$.000} & .585\,{\tiny$\pm$.004} & .573\,{\tiny$\pm$.037} & .647 \\
Agreement ST & .714\,{\tiny$\pm$.005} & .818\,{\tiny$\pm$.003} & .754\,{\tiny$\pm$.013} & .724\,{\tiny$\pm$.003} & .945\,{\tiny$\pm$.003} & .715\,{\tiny$\pm$.014} & .778 \\
LLM distill & .771\,{\tiny$\pm$.001} & .849\,{\tiny$\pm$.002} & .967\,{\tiny$\pm$.001} & .844\,{\tiny$\pm$.002} & .966\,{\tiny$\pm$.001} & .767\,{\tiny$\pm$.001} & .861 \\
\midrule
\multicolumn{8}{l}{\emph{Hybrid at 2{,}000 purchased labels} (Table~\ref{tab:paid-results}; LLM-distillation first stage)} \\
Hybrid (random) & .844\,{\tiny$\pm$.003} & .896\,{\tiny$\pm$.005} & .971\,{\tiny$\pm$.001} & .881\,{\tiny$\pm$.006} & .969\,{\tiny$\pm$.001} & .834\,{\tiny$\pm$.002} & .899 \\
Hybrid (BERT entropy) & .864\,{\tiny$\pm$.001} & .872\,{\tiny$\pm$.006} & .971\,{\tiny$\pm$.001} & .872\,{\tiny$\pm$.002} & .970\,{\tiny$\pm$.001} & .862\,{\tiny$\pm$.006} & .902 \\
Hybrid (JSD) & .871\,{\tiny$\pm$.002} & .878\,{\tiny$\pm$.005} & .972\,{\tiny$\pm$.000} & .882\,{\tiny$\pm$.004} & .971\,{\tiny$\pm$.001} & .865\,{\tiny$\pm$.001} & .906 \\
Hybrid (symmetric KL) & .869\,{\tiny$\pm$.002} & .875\,{\tiny$\pm$.008} & .972\,{\tiny$\pm$.001} & .878\,{\tiny$\pm$.002} & .970\,{\tiny$\pm$.001} & .860\,{\tiny$\pm$.002} & .904 \\
Hybrid (LLaMA entropy) & .848\,{\tiny$\pm$.002} & .903\,{\tiny$\pm$.002} & .972\,{\tiny$\pm$.002} & .899\,{\tiny$\pm$.003} & .969\,{\tiny$\pm$.002} & .839\,{\tiny$\pm$.004} & .905 \\
\bottomrule
\end{tabular}
\caption{Seed-level variability. Every adaptation run is repeated with seeds 42, 43, and 44; cells give the mean and standard deviation across those runs, and the final column the mean across the six transfers. HatE, HateX, and Dav abbreviate HatEval, HateXplain, and Davidson. The source model is a single checkpoint per source dataset and the teacher score is deterministic, so neither has seed variance. Two BERT-ST cells (HatEval-source transfers) report the unadapted source model, since no target example passed the confidence filter.}
\label{tab:app-seeds}
\end{table*}

\subsection{Prior-work adaptation baselines}
\label{sec:appendix-baselines}

We compare our zero-human-label policies against four established unsupervised domain adaptation baselines. All four use neither target labels nor the LLM signal; they are run in our harness on our splits with the same encoder (\texttt{bert-base-uncased}), class-weighted cross-entropy for classification, and seeds 42, 43, and 44.

\begin{table*}[b]
\centering
\small
\begin{tabular}{lrrrrrrr}
\toprule
 & \multicolumn{2}{c}{Davidson$\rightarrow$} & \multicolumn{2}{c}{HatEval$\rightarrow$} & \multicolumn{2}{c}{HateXplain$\rightarrow$} & \\
\cmidrule(lr){2-3}\cmidrule(lr){4-5}\cmidrule(lr){6-7}
Baseline & HatE & HateX & Dav & HateX & Dav & HatE & Mean \\
\midrule
Multi-source pooling$^{\ddagger}$ & \textbf{.745}\,{\tiny$\pm$.004} & .719\,{\tiny$\pm$.007} & .730\,{\tiny$\pm$.027} & .719\,{\tiny$\pm$.007} & \textbf{.730}\,{\tiny$\pm$.027} & \textbf{.745}\,{\tiny$\pm$.004} & \textbf{.731} \\
DAPT/TAPT & .738\,{\tiny$\pm$.011} & \textbf{.725}\,{\tiny$\pm$.006} & .825\,{\tiny$\pm$.035} & .676\,{\tiny$\pm$.038} & .617\,{\tiny$\pm$.035} & .688\,{\tiny$\pm$.013} & .711 \\
UDALM & \textbf{.745}\,{\tiny$\pm$.008} & .722\,{\tiny$\pm$.012} & .827\,{\tiny$\pm$.030} & .672\,{\tiny$\pm$.023} & .625\,{\tiny$\pm$.045} & .699\,{\tiny$\pm$.007} & .715 \\
Tri-training & .673\,{\tiny$\pm$.009} & .711\,{\tiny$\pm$.016} & \textbf{.855}\,{\tiny$\pm$.025} & \textbf{.738}\,{\tiny$\pm$.043} & .636\,{\tiny$\pm$.016} & .606\,{\tiny$\pm$.043} & .703 \\
\midrule
Agreement ST (ours) & .714\,{\tiny$\pm$.005} & .818\,{\tiny$\pm$.003} & .754\,{\tiny$\pm$.013} & .724\,{\tiny$\pm$.003} & .945\,{\tiny$\pm$.003} & .715\,{\tiny$\pm$.014} & .778 \\
LLM distillation (ours) & \textbf{.771}\,{\tiny$\pm$.001} & \textbf{.849}\,{\tiny$\pm$.002} & \textbf{.967}\,{\tiny$\pm$.001} & \textbf{.844}\,{\tiny$\pm$.002} & \textbf{.966}\,{\tiny$\pm$.001} & \textbf{.767}\,{\tiny$\pm$.001} & \textbf{.861} \\
\bottomrule
\end{tabular}
\caption{Prior-work unsupervised adaptation baselines, run in our harness on our splits with the same encoder, class-weighted cross-entropy for classification, and seeds (42, 43, and 44). Cells are target-test AUC, mean\,$\pm$\,standard deviation across seeds; bold marks the best value per column within each block. HatE, HateX, and Dav abbreviate HatEval, HateXplain, and Davidson. None of these baselines uses target labels or the LLM signal. The reference rows repeat, from Table~\ref{tab:free-results}, our two zero-human-label encoders: JSD-agreement self-training and LLM distillation. Both use no target labels but do use the LLM signal---for example selection and for pseudo-labels, respectively. $^{\ddagger}$ Multi-source pooling is leave-one-dataset-out: it trains on the two corpora other than the target, so its value depends only on the target and is identical for both transfers into that target.}
\label{tab:app-baselines}
\end{table*}

\paragraph{Multi-source pooling.}
Following \citet{antypas-camacho-collados-2023-robust}, we pool the training splits of all corpora except the target (leave-one-dataset-out) and evaluate on the target test split, giving roughly 6{,}500 to 12{,}000 training examples depending on the target. Because the training data depends only on which corpus is the target, this baseline yields one value per target, identical for both transfer directions into it. We also subsampled the pooled data to the size of the largest single source corpus. The subsampled variant retained most of the gain over source-only transfer (mean AUC 0.725, compared with 0.731 for full pooling).

\paragraph{Domain- and task-adaptive pretraining (DAPT/TAPT).}
We continue masked language modeling~(MLM) on the unlabeled target pool before supervised fine-tuning on the labeled source split \citep{gururangan-etal-2020-dont}. Because our pool is the task corpus itself, this is closest to task-adaptive pretraining in their taxonomy. We use masking probability 0.15 and MLM learning rate $5\times10^{-5}$, sweeping 3, 10, and 30 MLM epochs and selecting by unsupervised MLM validation loss; no target labels are used at any point, including model selection.

\paragraph{UDALM.}
Unsupervised Domain Adaptation through Language Modeling \citep{karouzos-etal-2021-udalm} keeps the MLM signal active during classifier training rather than discarding it after pretraining. A shared encoder is trained with two heads on the mixed objective
\begin{equation}
  \mathcal{L}
  = \lambda\,\mathcal{L}_{\mathrm{CE}}(\text{source})
  + (1-\lambda)\,\mathcal{L}_{\mathrm{MLM}}(\text{target}),
\end{equation}
where $\lambda$ is set to the source fraction of the mixed batch ($\lambda=1/9$). The first stage is the same continued MLM pretraining as DAPT. In the second stage, we train the classifier with both the source classification loss and the target MLM loss. Early stopping uses the same mixed criterion computed on held-out source and target slices, so no target labels enter at any stage.

\paragraph{Tri-training.}
We implement the agreement-based bootstrapping rule of \citet{zhou-li-2005}, re-evaluated as a strong neural domain-shift baseline by \citet{ruder-plank-2018}. Three classifiers are trained on bootstrap resamples of the labeled source set. In each of three rounds, each model is trained on the full labeled source set and on target-pool examples on which the other two models agree. At prediction time we average the three classifiers' probabilities; before any target round this ensemble already reaches 0.688 mean AUC, so about half of tri-training's gain over the source model comes from ensembling. We include it because it is the closest prior-work analogue of our own agreement arm: it pseudo-labels exactly where a committee agrees, but all committee members are source-trained BERTs, whereas our agreement arm pairs BERT with an LLM that brings target-relevant knowledge from outside the source corpus.

\section{OLMo Teachers on the Main Benchmarks}
\label{app:olmo}

With the OLMo teachers, we ran the zero-human-label policies (per-transfer results in Table~\ref{tab:app-olmo-zero}) and hybrid adaptation with LLM distillation followed by JSD, random, or teacher-entropy purchases (per-transfer results in Table~\ref{tab:app-olmo-paid}). BERT self-training and label-only adaptation do not use the teacher, so their values are shared with LLaMA-3. The original-HatEval evaluation uses LLaMA-3 only.

\begin{table*}[t]
\centering
\small
\resizebox{\textwidth}{!}{%
\begin{tabular}{lrrrrrl}
\toprule
Source \(\rightarrow\) target & Source BERT & Teacher & BERT ST & Agreement ST & LLM distill & Best zero-label policy \\
\midrule
\multicolumn{7}{l}{\textit{OLMo-7B-0724-Instruct teacher}} \\
Davidson \(\rightarrow\) HatEval & .704 & \textbf{.820} & .635 & .746 & \underline{.820} & OLMo (.820) \\
HatEval \(\rightarrow\) Davidson & .703 & \underline{.924} & .703\(^{\dagger}\) & .770 & \textbf{.966}\(^{*}\) & LLM distill (.966) \\
\midrule
Davidson \(\rightarrow\) HateXplain & .690 & \textbf{.875}\(^{*}\) & .715 & .807 & \underline{.845} & OLMo (.875) \\
HateXplain \(\rightarrow\) Davidson & .612 & .924 & .585 & \underline{.937} & \textbf{.965}\(^{*}\) & LLM distill (.965) \\
\midrule
HatEval \(\rightarrow\) HateXplain & .669 & \textbf{.875}\(^{*}\) & .669\(^{\dagger}\) & .725 & \underline{.840} & OLMo (.875) \\
HateXplain \(\rightarrow\) HatEval & .655 & \textbf{.820}\(^{*}\) & .573 & .719 & \underline{.814} & OLMo (.820) \\
\midrule
Mean & .672 & .873 & .647 & .784 & \underline{.875} & Per-transfer best (.887) \\
\midrule
\multicolumn{7}{l}{\textit{OLMo-1B-0724-Instruct teacher}} \\
Davidson \(\rightarrow\) HatEval & .704 & .605 & .635 & \textbf{.771} & \underline{.747}\(^{*}\) & Agreement ST (.771) \\
HatEval \(\rightarrow\) Davidson & \underline{.703} & .692 & .703\(^{\dagger}\) & .677 & \textbf{.806}\(^{*}\) & Tri-training (.855) \\
\midrule
Davidson \(\rightarrow\) HateXplain & .690 & .666 & .715 & \textbf{.771} & \underline{.728}\(^{*}\) & Agreement ST (.771) \\
HateXplain \(\rightarrow\) Davidson & .612 & .692 & .585 & \textbf{.872} & \underline{.865}\(^{*}\) & Agreement ST (.872) \\
\midrule
HatEval \(\rightarrow\) HateXplain & .669 & .666 & .669\(^{\dagger}\) & \underline{.720} & \textbf{.721}\(^{*}\) & Tri-training (.738) \\
HateXplain \(\rightarrow\) HatEval & .655 & .605 & .573 & \textbf{.754} & \underline{.731}\(^{*}\) & Agreement ST (.754) \\
\midrule
Mean & .672 & .654 & .647 & .761 & \underline{.766} & Per-transfer best (.793) \\
\bottomrule
\end{tabular}}
\caption{Zero-human-label results per transfer with the OLMo teachers (columns and marks as in Table~\ref{tab:free-results}; test in Appendix~\ref{sec:appendix}). Source BERT and BERT ST do not use the teacher, so they repeat the LLaMA-3 values. For OLMo-1B on HatEval\(\rightarrow\)Davidson and HatEval\(\rightarrow\)HateXplain, the strongest zero-human-label policy is tri-training (Table~\ref{tab:app-baselines}), which uses neither the teacher nor target labels.}
\label{tab:app-olmo-zero}
\end{table*}

\begin{table*}[b]
\centering
\small
\setlength{\tabcolsep}{4pt}
\resizebox{\textwidth}{!}{%
\begin{tabular}{llrrrrrr}
\toprule
Source \(\rightarrow\) target & \begin{tabular}[b]{@{}l@{}}Best\\zero-label\end{tabular} & AUC &
\begin{tabular}[b]{@{}r@{}}Teacher\\\(B_{\times}\)\end{tabular} & \begin{tabular}[b]{@{}r@{}}Best zero-label\\\(B_{\times}\)\end{tabular} & \begin{tabular}[b]{@{}r@{}}Hybrid\\@2,000\end{tabular} &
\begin{tabular}[b]{@{}r@{}}Gain\\@2,000\end{tabular} & Best @2,000 arm \\
\midrule
\multicolumn{8}{l}{\textit{OLMo-7B-0724-Instruct teacher}} \\
Davidson \(\rightarrow\) HatEval & OLMo & \underline{.820} & 200 & 200 & \textbf{.872} & \textbf{+.052} & LLM distill + JSD \\
HatEval \(\rightarrow\) Davidson & LLM distill & \underline{.966} & 200 & 200 & \textbf{.973} & \textbf{+.007} & LLM distill + JSD \\
\midrule
Davidson \(\rightarrow\) HateXplain & OLMo & \underline{.875} & 800 & 800 & \textbf{.898} & \textbf{+.023} & LLM distill + OLMo entropy \\
HateXplain \(\rightarrow\) Davidson & LLM distill & \underline{.965} & 200 & 200 & \textbf{.971} & \textbf{+.006} & LLM distill + OLMo entropy \\
\midrule
HatEval \(\rightarrow\) HateXplain & OLMo & \underline{.875} & 800 & 800 & \textbf{.893} & \textbf{+.018} & LLM distill + OLMo entropy \\
HateXplain \(\rightarrow\) HatEval & OLMo & \underline{.820} & 200 & 200 & \textbf{.865} & \textbf{+.045} & LLM distill + JSD \\
\midrule
Mean & Per-transfer best & \underline{.887} & 200--800 & 200--800 & \textbf{.912} & \textbf{+.025} & --- \\
\midrule
\multicolumn{8}{l}{\textit{OLMo-1B-0724-Instruct teacher}} \\
Davidson \(\rightarrow\) HatEval & Agreement ST & \underline{.771} & 200 & 400 & \textbf{.843} & \textbf{+.072} & LLM distill + random \\
HatEval \(\rightarrow\) Davidson & Tri-training & \underline{.855} & 200 & 200 & \textbf{.971} & \textbf{+.116} & LLM distill + JSD \\
\midrule
Davidson \(\rightarrow\) HateXplain & Agreement ST & \underline{.771} & 200 & 400 & \textbf{.876} & \textbf{+.105} & LLM distill + JSD \\
HateXplain \(\rightarrow\) Davidson & Agreement ST & \underline{.872} & 200 & 200 & \textbf{.967} & \textbf{+.095} & LLM distill + JSD \\
\midrule
HatEval \(\rightarrow\) HateXplain & Tri-training & \underline{.738} & 200 & 200 & \textbf{.876} & \textbf{+.138} & LLM distill + JSD \\
HateXplain \(\rightarrow\) HatEval & Agreement ST & \underline{.754} & 200 & 400 & \textbf{.822} & \textbf{+.068} & LLM distill + random \\
\midrule
Mean & Per-transfer best & \underline{.793} & 200 & 200--400 & \textbf{.892} & \textbf{+.099} & --- \\
\bottomrule
\end{tabular}}
\caption{Crossover budgets and AUC at 2,000 labels per transfer with the OLMo teachers (columns as in Table~\ref{tab:crossover-results}). The hybrid arms are LLM distillation followed by JSD, random, or teacher-entropy purchases. With OLMo-7B, the HatEval\(\rightarrow\)HateXplain crossing at 800 clears the reference by less than 0.001 AUC. The gain reaches at least 0.001 AUC at 1,000 labels. For OLMo-1B, tri-training is the reference on the two HatEval-source transfers (Table~\ref{tab:app-olmo-zero}).}
\label{tab:app-olmo-paid}
\end{table*}

\paragraph{Teachers.} LLaMA-3: Meta's original \texttt{Meta-Llama-3-8B-Instruct} checkpoint. OLMo-7B: \texttt{allenai/OLMo-7B-0724-Instruct-hf}, revision \texttt{aba3d33d}. OLMo-1B: \texttt{hamishivi/OLMo-1B-0724-Instruct-hf}, revision \texttt{152b0d72} (weights from the repository's safetensors conversion, pull request~1). OLMo-1B-0724-Instruct is not an official AllenAI release: it is the official OLMo-1B-0724 base model post-trained by H.~Ivison on the same instruction data and a cleaned version of the same preference data. The OLMo teachers use the same ten prompts, two demonstrations, and answer read-out as LLaMA-3 (Appendix~\ref{app:llama-prompts}, which lists the small differences), each with its own chat template. With the same prompting procedure, neither base OLMo-0724 model produced a prediction with confidence of at least 0.85 for any example in the benchmark adaptation pools.

\section{ToxiGen Decontamination and Results}
\label{app:toxigen}
\paragraph{ToxiGen target.} The three main benchmarks were public before the teachers were trained. To test adaptation after removing matches to OLMo's training data, we add the human-annotated subset of ToxiGen \citep{hartvigsen-etal-2022-toxigen} as a fourth target and use the same three source models. Following the LM Evaluation Harness \citep{biderman2024lessons}, we label a statement as hate when the sum of its toxicity scores for presumed human and AI authorship exceeds 5.5. We use the annotated test split for evaluation and form the adaptation pool from the annotated training split after removing the 120 texts that also appear in the test split.

\paragraph{Decontamination.} The OLMo teachers were pretrained on the public Dolma~1.7 corpus \citep{soldaini-etal-2024-dolma}. We adapted the overlap check of \citet{brown2020language} and fixed the rule before seeing any results. For each ToxiGen text, we search the infini-gram index \citep{liu2024infinigram} of Dolma~1.7 (3.4 billion documents) for an exact full-text match. If there is no match and the text is at least 13 words long, we search each window of 13 consecutive words. We exclude the text if any query returns a match. Full-text matches remove 32 test and 483 pool texts. The 13-word check removes a further 28 test and 345 pool texts. This leaves 8,012 of 8,840 pool texts and 880 of 940 test texts, with 43.0\% of the test texts labeled hate. Under the same rule, exact search finds no match for any of the remaining 8,892 texts in the OLMo teachers' Tulu~2 instruction data \citep{ivison2023camels} or either UltraFeedback preference set \citep{cui2024ultrafeedback}. LLaMA-3's training data are not public, so its exposure to ToxiGen remains unknown.

A sensitivity check ignores case and punctuation and also flags texts when at least 70\% of their 8-word windows match. It flags 16 pool texts (0.2\%), all short or generic phrases, and none of the 880 test texts. A looser 8-word rule would flag about half of the texts, mostly for ordinary English phrases. Other matching rules may detect additional overlap among the retained texts.

\paragraph{Benchmark overlap.} For the OLMo teachers, 55.6\%, 60.0\%, and 33.7\% of the Davidson, HatEval, and HateXplain test texts occur in Dolma~1.7 (whole text or a 13-word window), and 0.2--0.7\% occur in their post-training data, mostly inside instruction-tuning classification tasks that include the label. Teacher accuracy is not consistently higher on these texts, but they also differ from the rest in class balance and text type, so this neither shows nor excludes memorization.

\begin{table*}[t]
\centering
\small
\resizebox{\textwidth}{!}{%
\begin{tabular}{lrrrrrl}
\toprule
Source \(\rightarrow\) target & Source BERT & Teacher & BERT ST & Agreement ST & LLM distill & Best zero-label policy \\
\midrule
\multicolumn{7}{l}{\textit{LLaMA-3-8B-Instruct teacher}} \\
Davidson \(\rightarrow\) ToxiGen & .559 & \textbf{.915}\(^{*}\) & -- & .698 & \underline{.840} & LLaMA (.915) \\
HatEval \(\rightarrow\) ToxiGen & .739 & \textbf{.915}\(^{*}\) & -- & .739 & \underline{.833} & LLaMA (.915) \\
HateXplain \(\rightarrow\) ToxiGen & .597 & \textbf{.915}\(^{*}\) & -- & .626 & \underline{.846} & LLaMA (.915) \\
\midrule
Mean & .632 & \underline{.915} & -- & .688 & .840 & Per-transfer best (.915) \\
\midrule
\multicolumn{7}{l}{\textit{OLMo-7B-0724-Instruct teacher}} \\
Davidson \(\rightarrow\) ToxiGen & .559 & \textbf{.870}\(^{*}\) & -- & .715 & \underline{.840} & OLMo (.870) \\
HatEval \(\rightarrow\) ToxiGen & .739 & \textbf{.870}\(^{*}\) & -- & .701 & \underline{.830} & OLMo (.870) \\
HateXplain \(\rightarrow\) ToxiGen & .597 & \textbf{.870}\(^{*}\) & -- & .633 & \underline{.846} & OLMo (.870) \\
\midrule
Mean & .632 & \underline{.870} & -- & .683 & .839 & Per-transfer best (.870) \\
\midrule
\multicolumn{7}{l}{\textit{OLMo-1B-0724-Instruct teacher}} \\
Davidson \(\rightarrow\) ToxiGen & \underline{.559} & .357 & -- & \textbf{.623} & .358 & Agreement ST (.623) \\
HatEval \(\rightarrow\) ToxiGen & \textbf{.739} & .357 & -- & \underline{.562} & .371 & Source BERT (.739) \\
HateXplain \(\rightarrow\) ToxiGen & \textbf{.597} & .357 & -- & \underline{.579} & .385\(^{*}\) & Source BERT (.597) \\
\midrule
Mean & \underline{.632} & .357 & -- & .588 & .371 & Per-transfer best (.653) \\
\bottomrule
\end{tabular}}
\caption{Zero-human-label results on the ToxiGen target (decontaminated split; columns and marks as in Table~\ref{tab:free-results}). BERT ST was not run on ToxiGen. The source BERT is the best zero-label policy only with OLMo-1B. \protect\newline \(^{*}\)~Higher of the teacher and distilled BERT AUCs when the difference is significant after Holm correction across the three ToxiGen transfers (\(p<0.05\)).}
\label{tab:app-tox-zero}
\end{table*}

\begin{table*}[t]
\centering
\small
\resizebox{\textwidth}{!}{%
\begin{tabular}{llrrrrrr}
\toprule
Source \(\rightarrow\) target & Best zero-label & AUC & Teacher \(B_{\times}\) & Best zero-label \(B_{\times}\) & Hybrid @2,000 & Gain @2,000 & Best @2,000 arm \\
\midrule
\multicolumn{8}{l}{\textit{LLaMA-3-8B-Instruct teacher}} \\
Davidson \(\rightarrow\) ToxiGen & LLaMA & \underline{.915} & \(>\)2,000 & \(>\)2,000 & .859 & \(-\).056 & LLM distill + LLaMA entropy \\
HatEval \(\rightarrow\) ToxiGen & LLaMA & \underline{.915} & \(>\)2,000 & \(>\)2,000 & .859 & \(-\).056 & LLM distill + LLaMA entropy \\
HateXplain \(\rightarrow\) ToxiGen & LLaMA & \underline{.915} & \(>\)2,000 & \(>\)2,000 & .868 & \(-\).047 & LLM distill + LLaMA entropy \\
\midrule
Mean & Per-transfer best & \underline{.915} & \(>\)2,000 & \(>\)2,000 & .862 & \(-\).053 & --- \\
\midrule
\multicolumn{8}{l}{\textit{OLMo-7B-0724-Instruct teacher}} \\
Davidson \(\rightarrow\) ToxiGen & OLMo & \underline{.870} & \(>\)2,000 & \(>\)2,000 & .859 & \(-\).011 & LLM distill + OLMo entropy \\
HatEval \(\rightarrow\) ToxiGen & OLMo & \underline{.870} & \(>\)2,000 & \(>\)2,000 & .857 & \(-\).013 & LLM distill + OLMo entropy \\
HateXplain \(\rightarrow\) ToxiGen & OLMo & \underline{.870} & \(>\)2,000 & \(>\)2,000 & .868 & \(-\).002 & LLM distill + OLMo entropy \\
\midrule
Mean & Per-transfer best & \underline{.870} & \(>\)2,000 & \(>\)2,000 & .861 & \(-\).009 & --- \\
\midrule
\multicolumn{8}{l}{\textit{OLMo-1B-0724-Instruct teacher}} \\
Davidson \(\rightarrow\) ToxiGen & Agreement ST & \underline{.623} & 200 & \(>\)2,000 & .536 & \(-\).087 & LLM distill + random \\
HatEval \(\rightarrow\) ToxiGen & Source BERT & \underline{.739} & 200 & \(>\)2,000 & .537 & \(-\).202 & LLM distill + random \\
HateXplain \(\rightarrow\) ToxiGen & Source BERT & \underline{.597} & 200 & \(>\)2,000 & .577 & \(-\).020 & LLM distill + random \\
\midrule
Mean & Per-transfer best & \underline{.653} & 200 & \(>\)2,000 & .550 & \(-\).103 & --- \\
\bottomrule
\end{tabular}}
\caption{Paid adaptation on the ToxiGen target (columns as in Table~\ref{tab:crossover-results}). The hybrid arms are LLM distillation followed by JSD, random, or teacher-entropy purchases. \(>\)2,000: not reached within 2,000 labels.}
\label{tab:app-tox-paid}
\end{table*}

\paragraph{Results.} The distilled BERT classifiers scored below both strong teachers (Table~\ref{tab:app-tox-zero}). With LLaMA-3, mean AUC was 0.840 for distillation and 0.915 for the teacher (95\% bootstrap interval on 880 test texts: 0.896--0.933). The highest three-seed hybrid mean across sources and rules was 0.868 (Table~\ref{tab:app-tox-paid}). With OLMo-7B, the best hybrid per source averaged 0.861 at 2,000 labels, compared with 0.870 for the teacher (95\% bootstrap interval: 0.842--0.896). No hybrid arm surpassed either teacher within 2,000 labels. The strong teachers' pseudo-labels still help BERT: with random purchases, hybrid adaptation exceeds label-only adaptation at 2,000 labels by 0.055 (LLaMA-3) and 0.054 (OLMo-7B). OLMo-1B-Instruct scored below chance on each of its ten prompts (AUC 0.357 for their mean score), and 61\% of the pool passed its confidence filter. The best hybrid per source averaged 0.550 at 2,000 labels, below source BERT and label-only adaptation (0.800). A small labeled sample could reveal the teacher's weak performance before distillation.

\end{document}